\documentclass[runningheads]{llncs}

\usepackage{eccv}

\usepackage{eccvabbrv}

\usepackage{graphicx}
\usepackage{booktabs}

\usepackage[accsupp]{axessibility}  %

\usepackage[pagebackref=true,breaklinks=true,colorlinks,bookmarks=false]{hyperref}

\usepackage{orcidlink}
\usepackage{comment} %
\usepackage{balance}
\usepackage{arydshln}
\usepackage{multirow}
\usepackage{colortbl}
\usepackage{multicol}
\usepackage{graphicx}
\usepackage{wrapfig}
\usepackage{soul}

\definecolor{mypurple}{RGB}{0, 128, 255}
\definecolor{dexcolor}{RGB}{0, 117, 4}
\definecolor{honotatecolor}{RGB}{1, 0, 253}

\excludecomment{comments}

\excludecomment{remove}

\newcommand{\colorRef}[1]{\textcolor{black}{#1}} %
\newcommand{\reffig}[1]{\colorRef{Fig.~\ref{#1}}}

\newcommand{\refFig}[1]{\mbox{\colorRef{Figure~\ref{#1}}}}

\newcommand{\refTab}[1]{\mbox{\colorRef{Table~\ref{#1}}}}

\newcommand{\ccite}[1]{~\cite{#1}}

\definecolor{GreenColor}{rgb}{0.137,0.573,0.565}
\definecolor{OrangeColor}{rgb}{0.914,0.541,0.0.141}
\definecolor{PurpleColor}{rgb}{0.5,0,0.7}
\definecolor{BlueColor}{rgb}{0,0.725,0.949}
\definecolor{PinkColor}{rgb}{0.9843,0.19215,0.6}

\newcommand{\V}[1]{\mathbf{#1}} %
\newcommand{\R}{\rm I\!R}

\newcommand{\myparagraph}[1]{\noindent\textbf{#1:}}

\newcommand{\nameCOLOR}[1]{\textcolor{black}{#1}} %
\newcommand{\arctic}{\mbox{\nameCOLOR{ARCTIC}}\xspace}
\newcommand{\assembly}{\mbox{\nameCOLOR{AssemblyHands}}\xspace}

\newcommand{\methodnameSF}{\mbox{\nameCOLOR{ArcticNet-SF}}\xspace}
\newcommand{\arcticnet}{\mbox{\nameCOLOR{ArcticNet-SF}}\xspace}

\newcommand{\ambiguoushands}{\mbox{\nameCOLOR{AmbiguousHands}}\xspace}
\newcommand{\RWTH}{\mbox{\nameCOLOR{JointTransformer}}\xspace}

\newcommand{\uvhand}{\mbox{\nameCOLOR{UVHand}}\xspace}
\newcommand{\digit}{\mbox{\nameCOLOR{DIGIT}}\xspace}

\newcommand{\suppl}{\textcolor{black}{SupMat}\xspace}

\newcommand{\rgb}{RGB\xspace}

\newcommand{\threeD}{\xspace{3D}\xspace}

\newcommand{\sota}{{state-of-the-art}\xspace}

\newcommand{\groundtruth}{{ground-truth}\xspace}

\newcommand{\mano}{\mbox{MANO}\xspace}

\newcommand{\mytitle}[1]{\textbf{#1}.}
\newcommand{\TITLE}{
Benchmarks and Challenges in Pose Estimation for Egocentric Hand Interactions with Objects
}

\newcommand{\AHbase}{\mbox{\nameCOLOR{Base}}\xspace}
\newcommand{\AHthree}{\mbox{\nameCOLOR{FRDC}}\xspace}
\newcommand{\AHfour}{\mbox{\nameCOLOR{Phi-AI}}\xspace}
\newcommand{\AHone}{\mbox{\nameCOLOR{JHands}}\xspace}
\newcommand{\AHtwo}{\mbox{\nameCOLOR{PICO-AI}}\xspace}

\usepackage{eccv}

\usepackage{eccvabbrv}

\usepackage{graphicx}
\usepackage{booktabs}

\usepackage[accsupp]{axessibility}  %

\usepackage{hyperref}

\usepackage{orcidlink}

\begin{document}

\title{
\TITLE
} 

\titlerunning{Benchmarks and Challenges for Egocentric Hand Interactions}

\author{Zicong Fan$^{1,2*}$,~Takehiko Ohkawa$^{3*}$,~Linlin Yang$^{4*}$,\\~Nie Lin$^3$,
~Zhishan Zhou$^5$,~Shihao Zhou$^5$,~Jiajun Liang$^5$,\\~Zhong Gao$^6$,
~Xuanyang Zhang$^6$,~Xue Zhang$^7$,~Fei Li$^7$,~Zheng Liu$^8$,\\~Feng Lu$^8$,
~Karim Knaebel$^9$,~Bastian Leibe$^9$,~Jeongwan On$^{10}$,\\~Seungryul Baek$^{10}$,
~Aditya Prakash$^{11}$,~Saurabh Gupta$^{11}$,~Kun He$^{12}$,\\~Yoichi Sato$^3$,
~Otmar Hilliges$^1$,~Hyung Jin Chang$^{13}$,~Angela Yao$^{14}$}

\authorrunning{Z. Fan, T. Ohkawa, L. Yang et al.}

\institute{}

\newcommand\blfootnote[1]{%
  \begingroup
  \renewcommand\thefootnote{}\footnote{#1}%
  \addtocounter{footnote}{-1}%
  \endgroup
}

\maketitle
\begin{abstract}
We interact with the world with our hands and see it through our own (egocentric) perspective.
A holistic \threeD understanding of such interactions 
from egocentric views
is important for tasks in robotics, AR/VR, action recognition and motion generation. 
Accurately reconstructing such interactions in \threeD 
is challenging due to heavy occlusion, viewpoint bias, camera distortion, and motion blur from the head movement.
To this end, we designed the HANDS23 challenge based on the AssemblyHands and ARCTIC datasets with carefully designed training and testing splits. 
Based on the results of the top submitted methods and more recent baselines on the leaderboards, we perform a thorough analysis on \threeD hand(-object) reconstruction tasks. 
Our analysis demonstrates the effectiveness of addressing distortion specific to egocentric cameras, adopting high-capacity transformers to learn complex hand-object interactions, and fusing predictions from different views.
Our study further reveals challenging scenarios intractable with state-of-the-art methods, such as fast hand motion, object reconstruction from narrow egocentric views, and close contact between two hands and objects.
Our efforts will enrich the community's knowledge foundation and facilitate future hand studies on egocentric hand-object interactions.

\end{abstract}

\blfootnote{
$^*$Equal contribution,
$^1$ETH Z{\"u}rich,
$^2$Max Planck Institute for Intelligent Systems, T{\"u}bingen, Germany,
$^3$The University of Tokyo,
$^4$Communication University of China,
$^5$Jiiov,
$^6$Bytedance,
$^7$Fujitsu Research \& Development Center Co., Ltd.,
$^8$Beihang University,
$^9$RWTH Aachen University,
$^{10}$UNIST,
$^{11}$UIUC,
$^{12}$Meta,
$^{13}$University of Birmingham,
$^{14}$National University of Singapore.
}

\section{Introduction}
\label{sec:intro}

\begin{figure*}[t]
\centering
\includegraphics[width=0.7\hsize]{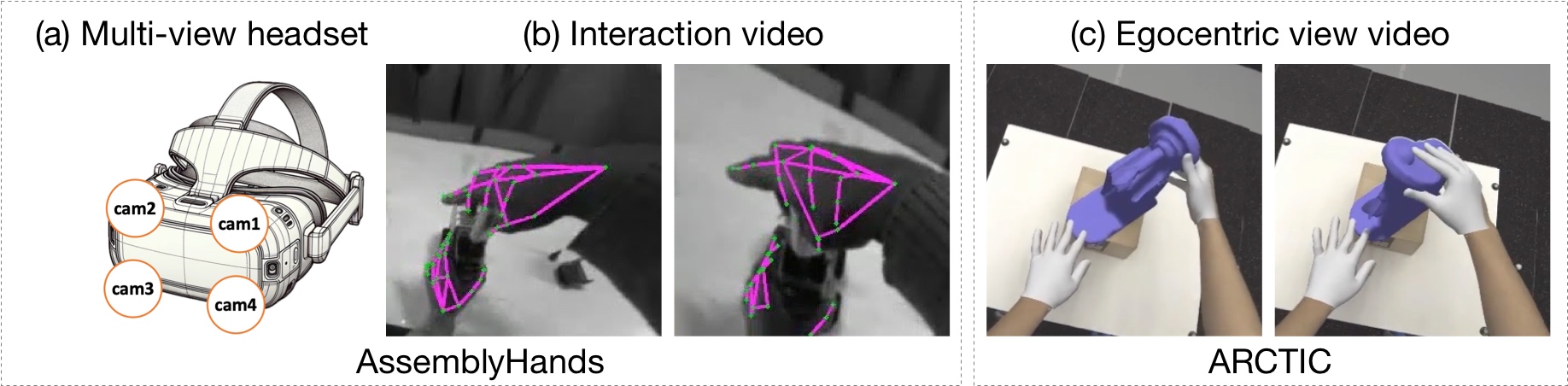}
\caption{\mytitle{Tasks in HANDS23}
In \assembly, from its multi-view headset (a), we estimate \threeD hand poses from images (b);  In \arctic, we  estimate the poses of two hands and articulated objects from an image (c).}
\label{fig:teaser}
\end{figure*}

We interact with the world with our hands
and see it through our eyes: %
we %
wake up and grab our phone to check the time; 
we %
use tools when assembling parts of a car; we open the microwave door to heat food, to name a few. 
An egocentric \threeD understanding of our hand interactions with objects 
will fundamentally impact areas such as robotics grasping~\cite{zhang2024artigrasp,christen2022dgrasp},
augmented and virtual reality~\cite{han:sigg22},
action recognition~\cite{garcia2018first,wen:arxiv23,chatterjee2024opening} and motion generation~\cite{zhang2024artigrasp,zhang2024graspxl}.

However, it is non-trivial to accurately reconstruct \threeD hands and or objects due to its high degree of freedoms~\cite{yang2021semihand,spurr2020eccv}, ambiguous texture~\cite{fan2021digit}, and heavy occlusions.
These challenges are intensified in an egocentric view~\cite{ohkawa:eccv22,ohkawa:access21}, particularly with object interactions, due to significant camera distortion, rapid and varied changes in %
viewpoint caused by head movements and hand-object occlusion.
To better understand these challenges, we introduce a public challenge in conjunction with ICCV 2023 ({\ie, HANDS23}) based on recent egocentric hand datasets, AssemblyHands~\cite{ohkawa2023assemblyhands} and ARCTIC~\cite{fan2023arctic} (see \refFig{fig:teaser}). These two datasets are large-scale, multi-view, and provide monochrome or RGB egocentric videos of the hands %
dexterously manipulating objects. 
Accordingly, we host two tasks: 1) egocentric 3D hand pose estimation from a single-view image based on AssemblyHands,  and 2) consistent motion reconstruction based on ARCTIC.

We introduce new HANDS23 methods and recent dataset leaderboard baselines that substantially outperform initial baselines for both tasks, setting new benchmarks for subsequent comparisons on the datasets. 
The two datasets are significantly larger and include a wider variety of bimanual manipulations compared to earlier datasets~\cite{hampali2020honnotate,dexycb}, enabling a more authentic assessment of real-world interactions.
With these benchmarks, we thoroughly analyze factors such as viewpoint, action types, hand position, model size, and object variations to determine their impact on \threeD hand(-object) reconstruction.

Our findings show the success of addressing the distortion of egocentric cameras with explicit perspective cropping or implicit learning for the distortion bias.
In addition, recent high-capacity vision transformers can learn complex hand-object interactions. 
Adaptive fusion techniques for multi-view predictions further boost performance.
We also analyze the remaining challenges that are still difficult to handle with the recent methods, \eg, fast hand motion, object reconstruction from narrow and moving views, and intricate interactions and close contact between two hands and objects.

To summarize, we contribute state-of-the-art baselines and gather the submitted methods for \assembly and \arctic to foster future research on egocentric hand-object interactions.
Furthermore, we thoroughly analyze the two benchmarks to provide insights for future directions in egocentric hand pose estimation and consistent motion reconstruction.

\section{Related work}
\myparagraph{\threeD hand pose estimation}
Reconstructing \threeD hand poses has a long history~\cite{erol:cviu07,ohkawa:ijcv23} ever since the important work led by Rehg and Kanade~\cite{Rehg:1994}. 
A large body of research in this area focuses on single-hand reconstruction~\cite{iqbal2018hand,Mueller2018ganerated,Spurr2018crossmodal,cai:eccv18,ge20193d,spurr2021peclr,spurr2020eccv,zimmermann2017iccv,Boukhayma2019,hasson2019obman,ziani2022tempclr,fan2021digit,simon2017hand,Zhang2019endtoend,tzionas2013directional,chen2023handavatar,fu2023deformer,dexycb,liu:cvpr24,duran2024hmp}.
For example, 
while a popular OpenPose library features 2D hand keypoints ~\cite{simon2017hand}, Zimmermann \etal~\cite{zimmermann2017iccv} initially extend to estimate \threeD hand poses using deep convolutional networks.
Ever since the release of InterHand2.6M~\cite{moon:eccv20} dataset, the community has increased focus on strongly interacting two hands reconstruction~\cite{kwon2021h2o,moon2023bringing,lee2023im2hands,li2022interacting,meng20223d,moon:eccv20,moon2024dataset,li2023renderih,ohkawa2023assemblyhands,guo2023handnerf}.
For example, Li~\etal~\cite{li2023renderih} and Moon~\etal~\cite{moon2024dataset} use relighting techniques to augment InterHand2.6M with more natural lighting and diverse backgrounds.

\myparagraph{Hand-object reconstruction} 
The holistic reconstruction of hands and objects have increased interest in the hand community in recent years~\cite{hasson2019obman,liu2021semi,grady2021contactopt,Hasson2020photometric,tekin2019ho,corona2020ganhand,zhou:cvpr20,hasson2021towards,tse2022collaborative,yang2021cpf,fan2023arctic,kwon2021h2o,dexycb}. 
Methods in this area mostly assume a given object template and jointly estimate the hand poses and the object rigid poses~\cite{hasson2019obman,liu2021semi,grady2021contactopt,Hasson2020photometric,tekin2019ho,corona2020ganhand,zhou:cvpr20,hasson2021towards,tse2022collaborative,yang2021cpf} or articulated poses~\cite{fan2023arctic}. 
For example, Cao \etal~\cite{cao2021handobject} fits object templates to in-the-wild interaction videos. 
Liu \etal~\cite{liu2021semi} introduce a semi-supervised learning framework via pseudo-groundtruth from temporal data to improve hand-object reconstruction.

More recent methods do not assume object templates for hand-object reconstruction~\cite{hasson2019obman,ye2022hand,chen2023gsdf,fan2024hold,ye2023vhoi,huang2022reconstructing,swamy2023showme}.
For example, Fan \etal~\cite{fan2024hold} introduced the first category-agnostic method that reconstructs an articulated hand and object jointly from a video. Nonetheless,%
our challenges and insights %
on hand-object occlusions and camera distortions are still applicable to these more challenging template-free reconstruction settings.

Public reports for the previous challenges (HANDS17~\cite{Yuan2018surveyHands} and HANDS19~\cite{armagan2020measuring}) 
distilled the insights from individual review papers and practical techniques into comprehensive summaries to enrich the community's knowledge base.
These past challenges use benchmarks which include depth-based hand pose estimation from egocentric views.
Instead of depth sensors, the HANDS23  benchmarks are based on affordable and widely applicable image sensors, \ie, RGB and monochrome images.
This paper further advances the analysis with unique insights, such as multi-view egocentric cameras, object reconstruction in contact, and modeling with recent transformers beyond conventional CNNs.

\section{HANDS23 challenge overview}

\label{sec:overview}
The workshop contains two hand-related \threeD reconstruction challenges in hand-object strongly interacting settings. 
In this section, we introduce the two challenges and their evaluation criteria.

\subsection{Workshop challenges}
\myparagraph{3D hand pose estimation in AssemblyHands}
As illustrated in \refFig{fig:teaser}, 
this task focuses on egocentric 3D hand pose estimation from a single-view image based on \assembly. The dataset provides multi-view captured videos of hand-object interaction while assembling and disassembling toy vehicles. 
In particular, it provides allocentric (fixed-view) and egocentric recordings and auxiliary cues like action, object, or context information for hand pose estimation. 
We refer readers to~\cite{ohkawa2023assemblyhands} for more dataset details.
The training, validation, and testing sets contain 383K, 32K, and 62K monochrome images, respectively, all captured from egocentric cameras.
During training, 3D hand keypoint coordinates, hand bounding boxes and camera intrinsic and extrinsic matrices for the four egocentric cameras attached to the headset are provided. 
The same information is provided during testing, minus the 3D keypoints.
Unlike~\cite{ohkawa2023assemblyhands}, 
given the availability of multi-view egocentric images, this task lets participants develop multi-view fusion based on the corresponding multi-view images.

\myparagraph{Consistent motion reconstruction in ARCTIC}
Given an \rgb image, the goal of this task is to estimate poses of hands and articulated objects to recover the \threeD surfaces of the interaction (see \refFig{fig:teaser}).
We refer readers to~\cite{fan2023arctic} for more details.
The \arctic dataset contains data of hands dexterously manipulating articulated objects and videos from 8$\times$ allocentric views and 1$\times$ egocentric views. 
The official splits of the ARCTIC dataset are used for training, validation, and testing.
There are two sub-tasks: allocentric task and egocentric task. In the former, only allocentric images can be used for training and evaluation. For the latter, all images from the training set can be used for training while only the egocentric view images are used during evaluation.

\subsection{Evaluation criteria}
\label{sec:eva}
\myparagraph{\assembly evaluation} 
We use MPJPE as an evaluation metric in millimeters, comparing the model predictions against the \groundtruth in world coordinates.
We provide the intrinsic and extrinsic of the egocentric cameras to construct submission results defined in the world coordinates.
Assuming that the human hand has a total ${N}_J$ joints, we denote 
wrist-relative coordinates of the prediction and ground-truth as $\hat{J} \in \mathbb{R}^{{N}_J \times 3}$ and $J \in \mathbb{R}^{{N}_J \times 3}$, respectively.
Given a joint visibility indicator $\gamma_{i}$ per joint $J_i$, we compute the Euclidean distance between predicted and ground-truth joints as $\frac{1}{\sum_{i=1}^{N_J} \gamma_i} \sum_{i=1}^{N_J} \gamma_i \left\| \hat{J}_i - J_i \right\|_2$.
The visibility indicator offers per-joint binary labels, representing whether the annotated keypoints are visible from the given egocentric view.

\myparagraph{\arctic evaluation} 
Since the original \arctic paper~\cite{fan2023arctic} has a heavy focus on the quality of hand-object contact in the reconstructed hand and object meshes, we use Contact Deviation (CDev) introduced in the \arctic dataset as the main metric for the competition. In particular, this metric measures the extent to which a hand vertex deviates from the supposed contact object vertex in the prediction. Concretely, 
suppose that for a given frame, $\{(\V{h}_i, \V{o}_i)\}_{i=1}^C$ are $C$ pairs of in-contact ($<3mm$ distance) hand-object vertices according to \groundtruth, and $\{(\hat{\V{h}}_i, \hat{\V{o}}_i)\}_{i=1}^C$ are the  predictions respectively. 
The CDev metric is the average distance between $\hat{\V{h}}_i$ and $\hat{\V{o}}_i$ in millimeters, $\frac{1}{C} \sum_{i=1}^C ||\hat{\V{h}}_i - \hat{\V{o}}_i||$.

For completeness, we report all metrics introduced in \arctic. 
In particular, the task requires the reconstructed meshes to have accurate hand-object contact (CDev) and smooth motion (ACC). Additionally, during articulation or when carrying an object, it is crucial that the vertices of the hand and object which are in contact maintain synchronized movement (MDev).
Moreover, we assess hand and object poses, alongside their relative movements, using metrics like MPJPE, AAE, Success Rate, and MRRPE. For detailed information, see \cite{fan2023arctic}.

\label{sec:res}

\section{Methods}
\label{sec:method}
This section presents the methods in the two challenges and other competing methods on the leaderboards.
Four methods outperform the baseline in both \assembly and \arctic.

\subsection{\assembly methods}\label{sec:ah_method}
Participants develop  methods that learn the mapping from egocentric images to 3D keypoints. 
The methods are categorized into: \textit{heatmap-based} and \textit{regression-based} approaches.
Given the presence of complex hand-object interactions in the egocentric scenes, high-capacity transformer models and attention mechanisms addressing occluded regions have been proposed as the backbone networks.
\refTab{tab:ah_summary} summarizes the methods based on the learning, preprocessing, multi-view fusion, and post-processing approaches.

\myparagraph{\AHbase} This method uses a \textit{heatmap-based} framework based on heatmaps~\cite{moon:eccv20} with 2.5D representations~\cite{iqbal2018hand} and a ResNet50~\cite{he:cvpr16} backbone.
The implementation can be found in~\cite{ohkawa:github23}.

\myparagraph{\AHone~\cite{zhou:cvpr24}} This method employs a \textit{regression-based} approach with simple MLP heads for regressing 2D keypoints, root-relative 3D keypoints, and the global root depth.
The regression training is empowered by a recent fast and strong vision transformer, Hiera~\cite{ryali:icml23}, pre-trained with masked auto-encoder~\cite{he:cvpr22}.
A multi-level feature fusion that concatenates the features of different layers is adopted for better feature extraction at different scales.
The method additionally uses other publicly available datasets for training, namely FreiHAND~\cite{zimmermann2017iccv}, DexYCB~\cite{dexycb}, and CompHand~\cite{chen:cvpr22}.
The implementation is available in~\cite{zhou:github24}.

\myparagraph{\AHtwo}
This method proposes a heatmap voting scheme in addition to the 2.5D heatmaps.
Due to their sparsity, the conventional heatmaps pose an imbalance problem between positive and negative samples in the loss function.
Hence, the proposed voting mechanism aims to spread the loss evenly across the entire heatmaps.
Given the initial guess of keypoints obtained from the heatmaps, the method defines a local region centered on the joint position and operates the soft-argmax within the region to obtain refined keypoint coordinates. 
This restricts the impact of background points, leading to more reliable optimization.
The training is facilitated by CNN-based RegNety320~\cite{radosavovic:cvpr20}.

\begin{table*}[t]
    \centering
    \caption{\textbf{Method and preprocessing summary in \assembly.}
    We summarize submitted methods in terms of learning methods, architecture, preprocessing, and multi-view fusion techniques. The tuple (views, phase) indicates the number of views used in either train or test time.
    }
    \label{tab:ah_summary}
\resizebox{1.0\linewidth}{!}{
\begin{tabular}{ccccc}
\toprule
Method & Learning methods                                                                      & Architecture                                                                                   & Preprocessing                                                                           & \begin{tabular}[c]{@{}c@{}}Multi-view fusion \\ (views, phase)\end{tabular}                  
\\ \hline
\addlinespace[3pt]
\AHbase       & 2.5D heatmaps~\cite{moon:eccv20}                                                               & ResNet50~\cite{he:cvpr16}                                                                               & -                                                                                                & Simple average (4, test)                                                                              \\
\addlinespace[3pt]
\hline
\addlinespace[3pt]
\AHone~\cite{zhou:cvpr24}        & Regression                                                                                     & Hiera~\cite{ryali:icml23}                                                                               & \begin{tabular}[c]{@{}c@{}}Warp perspective,\\ color jitter, random mask\end{tabular}            & \begin{tabular}[c]{@{}c@{}}Adaptive view selection\\ and average (2, test)\end{tabular}               \\ 
\addlinespace[3pt]
\hline
\addlinespace[3pt]
\AHtwo        & \begin{tabular}[c]{@{}c@{}}2.5D heatmaps~\cite{moon:eccv20}\\ Heatmap voting\end{tabular}      & RegNety320~\cite{radosavovic:cvpr20}                                                                    & \begin{tabular}[c]{@{}c@{}}Scale, rotate,\\ flip, tanslate\end{tabular}                          & \begin{tabular}[c]{@{}c@{}}Adaptive view selection\\ FTL~\cite{remelli:cvpr20} (2, train)\end{tabular} \\ 
\addlinespace[3pt]
\hline
\addlinespace[3pt]
\AHthree      & \begin{tabular}[c]{@{}c@{}}Regression\\ 2D heatmaps\end{tabular}                               & \begin{tabular}[c]{@{}c@{}}HandOccNet~\cite{park2022handoccnet}\\ with ConvNeXt~\cite{liu:cvpr22}\end{tabular} & \begin{tabular}[c]{@{}c@{}}Scale, rotate,\\ color jitter\end{tabular}                            & \begin{tabular}[c]{@{}c@{}}Weighted average\\ (4, test)\end{tabular}                                                                            \\ 
\addlinespace[3pt]
\hline
\addlinespace[3pt]
\AHfour       & \begin{tabular}[c]{@{}c@{}}2D heatmaps and \\ 3D location maps~\cite{zhou:cvpr20}\end{tabular} & ResNet50~\cite{he:cvpr16}                                                                               & \begin{tabular}[c]{@{}c@{}}Scale, rotate, translate, \\ color jitter, gaussian blur\end{tabular} & \begin{tabular}[c]{@{}c@{}}Weighted average\\ (4, test)\end{tabular}                                                                           \\
\addlinespace[3pt]
\bottomrule
\end{tabular}
}
\end{table*}

\myparagraph{\AHfour}
While following the heatmap-based approach, this method adapts MinimalHand~\cite{zhou:cvpr20} with the ResNet50 backbone, where 2D heatmaps and 3D location maps are regressed. 
Instead of selecting 3D keypoint coordinates from the location maps,
the proposed method modifies it by using heatmap values to weight 3D keypoint coordinates, achieving a more robust estimation. 
Moreover, the method adds a residual structured layer after the original three-tier cascade networks to refine the calculated location maps.
The method further applies the ensemble of final keypoint outputs combined with the \AHbase.

\myparagraph{\AHthree}
This method adopts a hybrid approach by combining \textit{regression} with \textit{heatmap} for training. 
HandOccNet~\cite{park2022handoccnet} is modified to regress 3D keypoint coordinates and integrated with an additional branch of 2D heatmap regression.
HandOccNet enriches feature extraction with spatial attention mechanisms for occluded regions, making it robust under hand-object occlusions.
The method further utilizes a stronger ConvNeXt~\cite{liu:cvpr22} backbone and feature fusion from the 2D keypoint regressor.

\myparagraph{Preprocessing of egocentric images}
Compared to conventional static camera setups, egocentric images exhibit unique properties and biases, such as distortion, head camera motion, and different color representations. 
Thus, it is vital to preprocess egocentric images to alleviate these effects during training.
Augmentation techniques are detailed in~\refTab{tab:ah_summary}.

The method \AHone addresses the distortion issue with a warp perspective operation to make the hands near the edge less stretched.
While AssemblyHands provides rectified images converted from fisheye cameras to a pinhole camera model, they often include excessively stretched areas near the edges. 
To address this, the method calculates a virtual camera and corresponding perspective transformation matrix based on the pixel coordinates of the crop and the camera parameters.
The generated crops can be found in the analysis of~\refFig{fig:crop}.

\myparagraph{Multi-view fusion}
Since AssemblyHands  offers multi-view egocentric videos,
participants can optionally use the constraint of multi-view geometry and fusion techniques during training or inference.

While \AHbase uses a simple average of predicted keypoints from all four camera views in the test time, 
\AHtwo proposes multi-view feature fusion during training using Feature Transform Layers (FTL)~\cite{remelli:cvpr20}.
This FTL training requires fusing two out of four views; thus, the method chooses the most suitable views for every frame.
In cases with multiple candidates, the Intersection over Union (IoU) is computed between hand boxes from per-view predictions and 2D keypoints from previous 3D predictions. The two views with the highest IoUs are selected for their superior prediction reliability.

The methods \AHone, \AHthree, and \AHfour apply adaptive fusion in predicted keypoints during testing.
The method \AHone computes the MPJPE with each other view and selects two results of views with the lowest MPJPE, excluding noisy predictions in the average.
If the MPJPE is lower than a threshold, the mean of the two results is calculated as the final result. 
Otherwise, the result with a lower PA-MPJPE with the predictions in the previous frame is chosen.
The methods \AHthree and \AHfour use a weighted average for each view prediction, assigning weights based on each view's validation performance.

\myparagraph{Postprocessing}
Several postprocessing techniques, including test-time augmentation, smoothing, and model ensembling are used to enhance inference outcomes. 
In particular, the method \AHone applies an offline smooth (Savitzky-Golay) filter on each video sequence.

\subsection{\arctic methods}

\refTab{tab:method_summary} summarizes the  details for each method in terms of the input image dimensions, image backbones, learning rate scheduling, the number of training epochs, batch size, and cropping strategies.

\begin{table}[t]
\caption{\textbf{Method and preprocessing summary in \arctic.}
We summarize baselines on \arctic in terms of input dimensions, image backbones, learning rate scheduling, training epochs, batch size and the cropping used for input.
$^*$Method trains 50 epochs for decoder and 36 for backbone.
$^+$Learning rate is 1e-7 to 1e-4 with linear warmup for first 5\% step, and 1e-4 to 1e-7 with cosine decay for rest.
}
\resizebox{1.0\linewidth}{!}{
\begin{tabular}{ccccccc}
\toprule
Method
& Input size                               & Backbones & Learning rate schedule                                         & Training epochs                                                               & Batch size & Cropping                                              \\
\hline
ArcticNet-SF~\cite{fan2023arctic}          & $224 \times 224$                         & ResNet50  & 1e-5                                            & \begin{tabular}[c]{@{}c@{}}allocentric: 20\\ egocentric: 50\end{tabular}      & 64         & object                                                \\
\hline
DIGIT~\cite{fan2021digit}           & $224 \times 224$ & HRNet-W32 & 1e-5                                            & \begin{tabular}[c]{@{}c@{}}allocentric: 20\\ egocentric: 50\end{tabular}      & 64         & object                        \\
\hline
AmbiguousHands        & $224 \times 224$ & ResNet50  & 1e-5                                            & \begin{tabular}[c]{@{}c@{}}allocentric: 20\\ egocentric: 100\end{tabular}     & 32         & \begin{tabular}[c]{@{}c@{}}hand\\ object\end{tabular} \\
\hline
\uvhand                & $384 \times 384$                         & Swin-L  & \begin{tabular}[c]{@{}c@{}}2e-4 (backbone) \\1e-7 (others) \end{tabular} & \begin{tabular}[c]{@{}c@{}}allocentric: N/A \\ egocentric: 50/36$^*$\end{tabular} & 48         & object        \\     \hline
\RWTH~\cite{kknaebel2023transformer}              & $224 \times 224$                         & ViT-G & 1e-7/1e-5$^+$ & \begin{tabular}[c]{@{}c@{}}allocentric: 20  \\ egocentric: 100 \end{tabular} & 64         & object       \\  
\bottomrule
\end{tabular}
}
\label{tab:method_summary}
\end{table}

\myparagraph{Preliminary}
All methods below are regression-based and predict two-hand \mano\ccite{mano} parameters $\V{\Theta} = \{ \V{\theta}, \V{\beta} \}$ and articulated object parameters $\V{\Omega}$. 
In particular, with the MANO pose and shape parameters $\V{\theta}, \V{\beta}$, the \mano model $\mathcal{H}$ returns a mesh with vertices via $\mathcal{H}(\V{\theta}, \V{\beta}) \in \R^{778 \times 3}$.
\threeD joints are obtained via a linear regressor. 
The articulated object model $\mathcal{O}$ was introduced in \arctic to provide an articulated mesh with vertices via $\mathcal{O}(\V{\Omega}) \in \R^{V \times 3}$, where $\V{\Omega} \in \R^7$ contains the global orientation, global translation, and object articulation.

\myparagraph{\methodnameSF~\cite{fan2023arctic}} Introduced in \arctic, it is a single-frame baseline. It first extracts an image feature vector from the input image; then, it regresses hand and object parameters with simple MLPs.
The hand and object meshes can then be extracted via $\mathcal{H}(\cdot)$ and $\mathcal{O}(\cdot)$. 
For more details, see~\cite{fan2023arctic}.

\myparagraph{\RWTH~\cite{kknaebel2023transformer}} 
\RWTH enhances \methodnameSF by integrating a transformer decoder instead of the MLP regressors for hand and object parameter estimation.
The decoder employs learned queries for the angle of each joint, the shape and translation of each hand, and the translation, rotation, and articulation of the object.
It alternates between self-attention between queries and cross-attention of queries to the elements of the backbone feature map, followed by linear layers that regress the final parameters.
Specifically, separate linear layers are dedicated to regressing joint angles, hand shape, hand translation, object translation, object orientation, and object articulation.
The best model uses a ViT-G~\cite{zhai2022scaling}  backbone with frozen DINOv2 weights~\cite{oquab2023dinov2}.

\myparagraph{\ambiguoushands~\cite{Prakash2023Hands}} The method addresses scale ambiguity, resulting from bounding box cropping in data augmentation and camera intrinsics, by employing positional encoding of these elements to mitigate scale issues. This leads to improved spatial alignment. Subsequently, the approach enhances network visibility by integrating local features through distinct hand and object crops. They follow the general approach of \methodnameSF to regress hand/object parameters.

\myparagraph{\uvhand} Since \methodnameSF only leverages a global feature vector to estimate hand and object parameters, the image features lack local context. To address this, \uvhand leverages Swin-L transformer~\cite{liu2021swin} to extract image features. They then further leverage Deformable DETR~\cite{zhu:iclr21} to encode the multiple-scale feature maps. The encoded feature maps are then aggregated via self- and cross-attention before regresssing hand and object parameters.

\myparagraph{\digit~\cite{fan2021digit}} The method was introduced to estimate strongly interacting hands in~\cite{fan2021digit}. Since \methodnameSF is sensitive when the hands are interacting with objects,  \digit was extended to the \arctic setting. 
Given an image, it first estimates hand part-wise segmentation masks and object masks.
The mask predictions are fused with the image features to perform parameter estimation.

\myparagraph{Implementation details} 
\refTab{tab:method_summary} shows that all methods use the default cropping as in \arctic to crop around the object, while \ambiguoushands performs three crops (around two hands and the object).
\digit uses the HRNet-W32 backbone~\cite{sun2019deep} and trains with a batch size of 64 with the same learning rate for all iterations.
\uvhand takes as input a $384\times 384$ image cropped around the object and encodes it with the Swin-L transformer~\cite{liu2021swin} backbone. It was trained with a batch size of 48 with a learning rate of 2e-4 for the backbone and 1e-7 for other weights. 
Due to computational cost, they train 50 epochs for the decoder and 36 for the backbone.
\RWTH uses ViT-G~\cite{zhai2022scaling} backbone with frozen DINOv2~\cite{oquab2023dinov2} weights to train with a batch size of 64. It performs a linear warmup from 1e-7 to 1e-4 in the first 5\% steps and uses cosine decay from 1e-4 to 1e-7 for the rest of the steps.

\section{Results and analysis}

\begin{table*}[t]
    \caption{\textbf{Method performance in \assembly.}
    We compare \assembly method performance on egocentric test data.
    We show the final MPJPE on the test set as the metrics (lower better). 
    We also provide detailed evaluations, regarding the varying distances of hand position from the image center and different verb action categories.
    The hand distance is computed by the distance from the image center to the hand center position per image, and averaged over the lower two views of the headset.
    Verb classes of ``attempt to X'' are merged to ``X'' for simplicity.
    The higher and lower three verbs are color-coded in red and blue, respectively.
    }
    \label{tab:assembly}
    \resizebox{1.0\linewidth}{!}{
\begin{tabular}{cccccccccccc}
\toprule
\multicolumn{1}{l|}{}         & \multicolumn{1}{l|}{}         & \multicolumn{3}{c|}{\textbf{Hand distance (px)}}                                 & \multicolumn{7}{c||}{\textbf{Verb class}}                                                                                                                                                                                                                      \\
\multicolumn{1}{l|}{Method}     & \multicolumn{1}{c|}{Score}    & -200     & 200-250 & \multicolumn{1}{c|}{250-}                                   & clap                          & inspect                       & pass                          & pick up                       & position                      & \begin{tabular}[c]{@{}c@{}}position \\ screw on\end{tabular} & \multicolumn{1}{c||}{pull}                          \\ \hline
\multicolumn{1}{l|}{\AHbase}  & \multicolumn{1}{c|}{20.69}    & 20.31    & 21.97   & \multicolumn{1}{c|}{24.85}                                  & 19.88                         & \cellcolor[HTML]{F4CCCC}22.89 & 22.48                         & 21.85                         & 22.65                         & 21.62                                                        & \multicolumn{1}{c||}{18.74}                         \\
\multicolumn{1}{l|}{\AHone}   & \multicolumn{1}{c|}{12.21}    & 12.35    & 11.98   & \multicolumn{1}{c|}{13.72}                                  & \cellcolor[HTML]{C9DAF8}10.65 & \cellcolor[HTML]{F4CCCC}16.27 & 12.86                         & 13.67                         & \cellcolor[HTML]{F4CCCC}14.58 & 12.8                                                         & \multicolumn{1}{c||}{11.06}                         \\
\multicolumn{1}{l|}{\AHtwo}   & \multicolumn{1}{c|}{12.46}    & 12.51    & 11.62   & \multicolumn{1}{c|}{12.95}                                  & 12.98                         & \cellcolor[HTML]{F4CCCC}15.3  & \cellcolor[HTML]{C9DAF8}11.37 & 13.2                          & 13.18                         & 11.39                                                        & \multicolumn{1}{c||}{15.13}                         \\
\multicolumn{1}{l|}{\AHthree} & \multicolumn{1}{c|}{16.48}    & 16.39    & 15.89   & \multicolumn{1}{c|}{18.69}                                  & 15.24                         & \cellcolor[HTML]{F4CCCC}21.33 & 17.86                         & 18.26                         & \cellcolor[HTML]{F4CCCC}19.21 & 18.03                                                        & \multicolumn{1}{c||}{\cellcolor[HTML]{C9DAF8}12.83} \\
\multicolumn{1}{l|}{\AHfour}  & \multicolumn{1}{c|}{17.26}    & 17.24    & 15.81   & \multicolumn{1}{c|}{19.51}                                  & \cellcolor[HTML]{F4CCCC}19.86 & \cellcolor[HTML]{F4CCCC}20.93 & 17.91                         & 19.01                         & 19.7                          & 19.35                                                        & \multicolumn{1}{c||}{17.3}                          \\
\addlinespace[2pt]
\midrule
                              & \multicolumn{11}{||c}{\textbf{Verb class (continue)}}                                                                                                                                                                                                                                                                                            %
                              \\
\multicolumn{1}{l}{}         & \multicolumn{1}{||c}{push}                          & put down & remove  & \begin{tabular}[c]{@{}c@{}}remove\\ screw from\end{tabular} & rotate                        & screw                         & tilt down                     & tilt up                       & unscrew                       & none                                                         &                               \\ \cline{1-11}
\multicolumn{1}{l}{\AHbase}  & \multicolumn{1}{||c}{19.29}                         & 20.26    & 19.99   & \cellcolor[HTML]{C9DAF8}16.47                               & \cellcolor[HTML]{F4CCCC}22.71 & \cellcolor[HTML]{F4CCCC}22.95 & \cellcolor[HTML]{C9DAF8}13.12 & \cellcolor[HTML]{C9DAF8}15.11 & 20.78                         & 19.82                                                        &                               \\ 
\multicolumn{1}{l}{\AHone}   & \multicolumn{1}{||c}{13.96}                         & 13.72    & 13.11   & 11.72                                                       & 12.26                         & \cellcolor[HTML]{F4CCCC}14.11 & \cellcolor[HTML]{C9DAF8}9.61  & \cellcolor[HTML]{C9DAF8}9.92  & 12.25                         & 10.99                                                        &                               \\
\multicolumn{1}{l}{\AHtwo}   & \multicolumn{1}{||c}{12.29}                         & 13.41    & 12.83   & \cellcolor[HTML]{C9DAF8}9.99                                & \cellcolor[HTML]{F4CCCC}13.7  & \cellcolor[HTML]{F4CCCC}13.72 & \cellcolor[HTML]{C9DAF8}10.44 & 11.56                         & 12.87                         & 11.71                                                        &                               \\
\multicolumn{1}{l}{\AHthree} &  \multicolumn{1}{||c}{\cellcolor[HTML]{F4CCCC}19.52} & 17.87    & 18.55   & 14.47                                                       & 16.44                         & 18.81                         & \cellcolor[HTML]{C9DAF8}14.03 & \cellcolor[HTML]{C9DAF8}13.37 & 16.41                         & 15.01                                                        &                               \\
\multicolumn{1}{l}{\AHfour}  &  \multicolumn{1}{||c}{19.12}                         & 18.29    & 18.18   & \cellcolor[HTML]{C9DAF8}13.95                               & 18.35                         & \cellcolor[HTML]{F4CCCC}19.78 & \cellcolor[HTML]{C9DAF8}13.13 & \cellcolor[HTML]{C9DAF8}15.36 & 17.29                         & 15.8                                                         &                              \\
\bottomrule
\end{tabular}
}
\end{table*}

\begin{table}[t]
\caption{
\mytitle{Method performance in \arctic}
We compare performance in both allocentric (top half) and egocentric (bottom half) views. 
We evaluate using metrics for contact and relative position (measuring hand-object contact and prediction of relative root position), motion (assessing temporally-consistent contact and smoothness), and hand and object metrics (indicating root-relative reconstruction error). 
We use the CDev score as the main metric for this competition.
We denote left and right hands as $l$ and $r$, and the object as $o$. 
}
\resizebox{1.0\linewidth}{!}{
\begin{tabular}{cc|cc|cc|c|cc}
\toprule
&               & \multicolumn{2}{|c|}{Contact and Relative Positions} & \multicolumn{2}{|c|}{Motion} & Hand         & \multicolumn{2}{c}{Object} \\
\hline       
\multicolumn{2}{c}{Method}         & CDev [$mm$] $\downarrow$& MRRPE$_{rl/ro}$ [$mm$] $\downarrow$ & MDev [$mm$] $\downarrow$ & ACC$_{h/o}$ [$m/{s^2}$] $\downarrow$ & MPJPE [$mm$] $\downarrow$ & AAE [$^\circ$] $\downarrow$ & Success Rate [$\%$] $\uparrow$\\
\hline
\parbox[t]{0mm}{\multirow{5}{*}{\rotatebox[origin=c]{90}{Allocentric}}}\hspace{4mm} & ArcticNet-SF   & 41.56                  & 52.39/37.47               & 10.40      & 5.72/7.57     & 21.45        & 5.37     & 71.39           \\
& \digit   & 34.92                  & 44.19/35.43               & \textbf{8.37}      & \textbf{4.86}/6.63     & 17.92        & 5.24     & 76.52           \\
& \uvhand         & 64.15                  & 84.68/70.31               & 14.12      & 7.05/12.04    & 40.99        & 12.36    & 31.47           \\
& \ambiguoushands & 33.25                  & 45.78/34.56               & 10.12      & 6.37/6.40     & 18.02        & 4.64     & 81.94           \\
& \RWTH       & \textbf{27.97}                  & \textbf{36.17/28.18}               & 8.93       & 6.08/\textbf{5.79}     & \textbf{17.12}        & \textbf{3.95}     & \textbf{89.79}           \\
\midrule
\parbox[t]{0mm}{\multirow{5}{*}{\rotatebox[origin=c]{90}{Egocentric}}}\hspace{4mm} & ArcticNet-SF   & 44.71                  & 28.31/36.16               & 11.80      & 5.03/9.15     & 19.18        & 6.39     & 53.89      \\
& \digit   & 41.31                  & 25.49/32.61               & \textbf{9.48}      & 4.01/8.32     & 16.74        & 6.60     & 53.33           \\
& \uvhand         & 40.43                  & 40.93/36.88               & 9.96       & 5.32/8.33     & 24.53        & 7.32     & 57.28           \\
& \ambiguoushands & 35.93                  & \textbf{23.07}/27.53               & 9.51       & \textbf{3.95/6.76}     & \textbf{16.26}        & 4.86     & 68.36           \\
& \RWTH       & \textbf{32.56}                  & 26.07/\textbf{26.22}               & 11.34      & 5.52/8.68     & 16.33        & \textbf{4.44}     & \textbf{74.07}           \\
\bottomrule
\end{tabular}
}
\label{tab:arctic_benchmark}
\end{table}

\subsection{Results}
Here, we benchmark results of valid submissions for \sota comparison in \assembly and \arctic and other more recent baselines. In particular, for \assembly, we report egocentric hand pose estimation results.
For \arctic, we report results for the allocentric and egocentric test sets.

\myparagraph{\assembly benchmark}
\refTab{tab:assembly} shows the final test scores on the AssemblyHands dataset. 
The methods in the table exceed the baseline (\AHbase) with a test score of 20.69 MPJPE.
Notably, the methods \AHone and \AHtwo achieve a nearly 40~\% reduction over the baseline. 
The methods \AHthree and \AHfour improve the test score by 20.3~\% and 16.5~\% against the baseline, respectively.

\myparagraph{\arctic benchmark}
\refTab{tab:arctic_benchmark} presents the comparative performance of methods in the \arctic dataset, where \arcticnet serves as the initial benchmark. The majority surpass \arcticnet in both allocentric and egocentric views, except for \uvhand, which underperforms due to incomplete training. In the egocentric view, \ambiguoushands excels in creating smooth, consistent mesh motions (refer to MDev and ACC$_{h/o}$ metrics). Notably, \RWTH stands out by significantly lowering CDev errors by $32.7\%$ in allocentric and $27.2\%$ in egocentric settings compared to the baseline.

\subsection{\assembly analysis}
We provide  analysis, regarding action-wise evaluation, distortion effect in training, and the effect of multi-view fusion. See \suppl for addtional results.

\begin{figure*}[t]
\centering
\includegraphics[width=1\hsize]{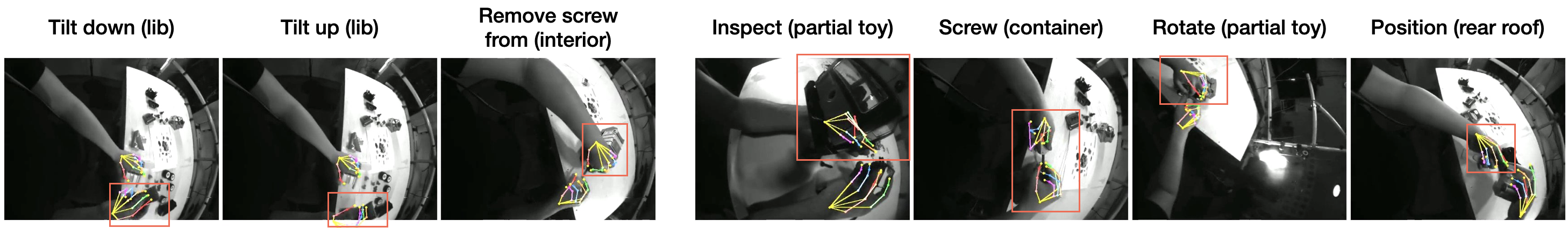}
\caption{\textbf{Qualitative results per action in AssemblyHands}.
We show \AHbase results with ``verb (noun)'' actions.
The left three figures are lower error situations while the right four ones are failure cases. 
The red boxes denote the area where the action occurs.
}
\label{fig:action}
\end{figure*}

\myparagraph{Action-wise evaluation}\label{sec:ah_results}
To analyze errors related to hand-object occlusions and interactions, 
we show pose evaluation according to fine verb action classes in \refTab{tab:assembly}.
We use the verb classes annotated by Assembly101~\cite{sener2022assembly101}, spanning every few seconds in a video.
\refFig{fig:action} shows qualitative results of representative verb classes with the top and bottom error cases. 

We observe that the performance varies among different verb actions. 
The verbs ``tilt down/up'' and ``remove screw from'' exhibit lower errors among the submitted methods, because hands are less occluded and their movement is relatively stable.
The ``tilt'' action holds a small part of the toy and turns it around alternately, leading to less overlap between the hand and the object (lib). The ``remove screw from'' action takes a screw out from the toy vehicle by their hand where observed hand poses do not change drastically.

Higher error classes, such as ``inspect'', ``screw'', ``rotate'', and ``position'', contain heavy occlusions, fast hand motion, complex two hands and object interactions.
The ``inspect'' action brings the toy close to the human eyes where the toy occupies a large portion of the image causing heavy object occlusions.
The ``screw'' action involves intricate fingertip movements to rotate the screwdriver quickly.
The ``rotate'' and ``position'' actions are performed so that the two hands and the object interact in close contact, which complicates the estimation.
We observe that the top two methods \AHone and \AHtwo significantly correct the results of these higher error actions compared to the other submitted methods. 

\begin{figure*}[t]
  \centering
  \begin{minipage}{0.64\linewidth}
    \centering
    \includegraphics[width=0.95\hsize]{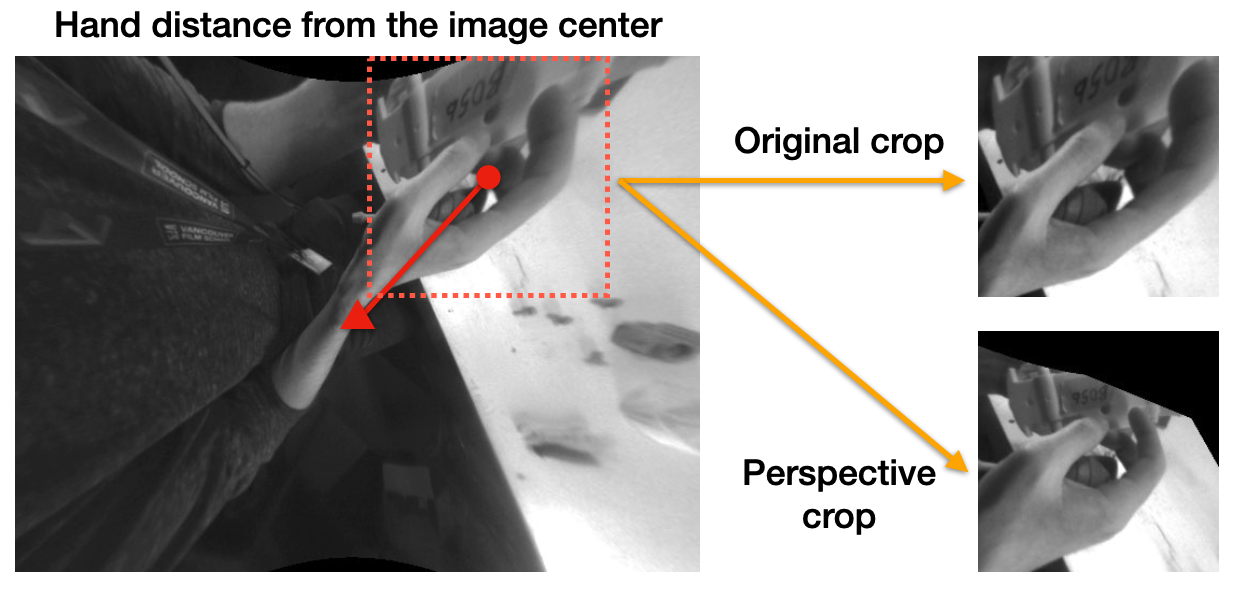}
    \caption{\textbf{Effect of distortion in AssemblyHands}.
The officially released images in the dataset have highly stretched areas near the edges (original crop). The method \AHone uses a perspective crop with a virtual camera to correct this distortion.
    }
    \label{fig:crop}
  \end{minipage}
  \hfill %
  \begin{minipage}{0.32\linewidth}
    \centering
    \captionof{table}{\textbf{Multi-view fusion in AssemblyHands.}
    We use the \AHbase result to show performance before and after fusion. 
    Missing instances per view are denoted as ``Miss(\%)''.
    }
    \begin{tabular}{lcc}
    View     & \multicolumn{1}{l}{MPJPE} & \multicolumn{1}{l}{Miss(\%)} \\ \hline
    cam1     & 37.97                     & 70.8                            \\
    cam2     & 25.71                     & 88.3                            \\
    cam3     & 22.19                     & 0.92                            \\
    cam4     & 22.29                     & 0.74                            \\ \cdashline{1-3}
    cam3+4   & 21.52                     & 0.08                            \\
    all four & 20.69                     & 0                              
    \end{tabular}    
    \label{tab:mv_fusion}
  \end{minipage}
\end{figure*}

\myparagraph{Bias of hand position in an image}
Hands near image edges are highly distorted due to the fish-eye cameras.
Directly using these noisy images in training will degrade  performance~\cite{zhou:arxiv23}; thus, some methods  create new crops with less distortion, select training instances, or adaptively fuse predictions during the inference.
Specifically, \refFig{fig:crop} shows that the method \AHone reformulates the perspective during cropping and creates less-distorted (perspective) crops.

To study this effect in the final performance, we split the evaluation instances into classes with different 2D distances between the hand center and the image center in \refTab{tab:assembly}.
Higher distances (250- pixels) indicate closer hand crops to image edges.
The method, \AHbase, without any training instance selection and distortion correction, has higher error as the crops approach image edges (20.31 $\rightarrow$ 24.85).
In contrast, the newly proposed methods are more robust and have a lower error, particularly in the 200-250 range.
We observe that the ranges 200-250 and 250- occupy 10\% and 5\% of the test images, respectively, thus the improvement in the 200-250 range helps the lowering of the overall score.

\myparagraph{Effect of multi-view fusion}
The multi-view egocentric camera setup is unique to the dataset.
We show the statistics and performance of multi-view fusion in \refTab{tab:mv_fusion}.
Note that \refTab{tab:assembly} shows the final results after multi-view fusion.

We found that samples captured from the lower cameras (cam3 and cam4; see \refTab{fig:teaser} for the layout) are numerous (fewer missing samples) and their errors are lower as they are faced toward the area occurring hand interactions.
Conversely, the samples from cam1 and cam2 are fewer and unbalanced as 
their cameras often fail to capture hands due to the camera layout.
For instance, cam1 (top-left) tends to capture more hand region than cam2 as the participants are mostly right-handed and bring up the object with the right hand, which can be better observed from cam1.
Given this uneven sample distribution, the proposed adaptive view selection methods in either training or testing are essential to perform effective multi-view fusion, and outperform the \AHbase's test-time average using all views all the time (see \textbf{Multi-view fusion} in Section~\ref{sec:ah_method}).

We further study the performance gain before and after multi-view fusion using \AHbase's results.
While per-view performance achieves 22.19 and 22.29 in cam3 and cam4, respectively, their fused results with simple average reduce the error to 21.52. 
Merging all four views has shown to be more effective than two-view fusion (20.69 \vs 21.52), indicating a 6.5\% reduction compared to the single camera setup (cam3).
This suggests predictions from the top views (cam1 and cam2) are informative in averaging even when they are prone to be erroneous.

\subsection{\arctic analysis}
\begin{figure}[t]
  \centering
 \includegraphics[width=0.8\textwidth]{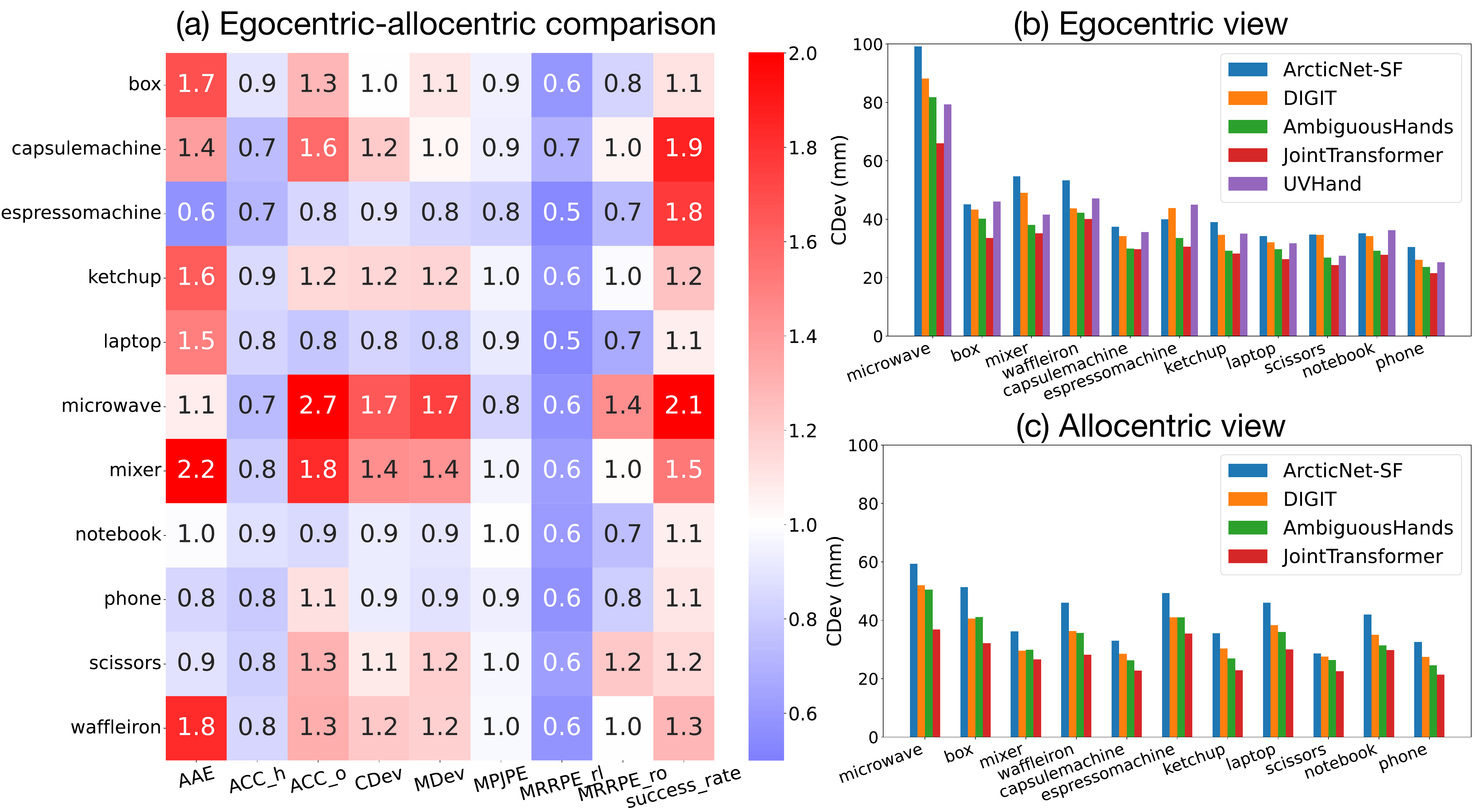}
    \caption{\mytitle{Performance comparison: Egocentric vs Allocentric} 
    (a) Comparative difficulty ratio of egocentric to allocentric views. 
    (b) Egocentric view performance by method across objects.
    (c) Allocentric view performance by method across objects.
    }
\label{fig:ego_allo_ratio}
\end{figure}

\myparagraph{Egocentric-allocentric comparison}
\refFig{fig:ego_allo_ratio}a compares the performance between egocentric and allocentric views. 
In particular, we compute a ratio between the metric values of the egocentric and allocentric view to %
measure the extent of difficulty for the egocentric view compared to the allocentric view.
Since success rate is a metric whose value is positively correlated to performance, we take its reciprocal ratio.
We average the ratios across methods and actions.

We observe that hand pose-related metrics such as MPJPE and ACC$_h$ are less than 1.0 on average (see blue color cells), meaning the the egocentric view is easier than allocentric view. 
This is because most allocentric cameras in \arctic are meters away from the subject while the egocentric camera is often close-up, offering higher hand visibility. 
Relative translation metrics between hand and object such as MRRPE$_{rl}$ are also easier in the egocentric view because estimating translation is more difficult from further cameras.

\begin{figure}[t]
  \centering
 \includegraphics[width=0.8\textwidth]{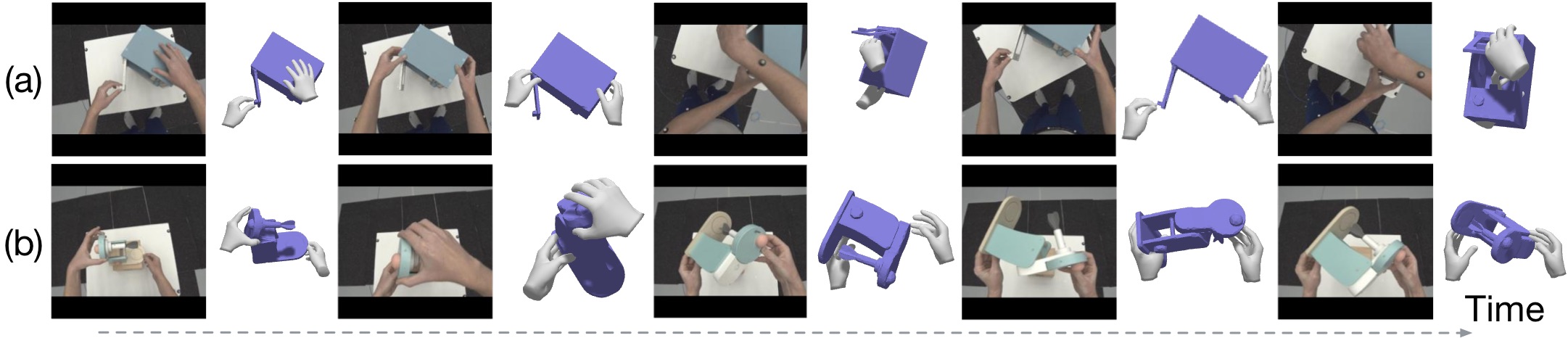}
    \caption{\mytitle{Egocentric reconstruction by top method in \arctic}
    In the egocentric view, object reconstruction struggle when the object is partially observed on the image boundaries, as well as when heavy hand/arm occlusion occurs.
    }
\label{fig:failure_cases}
\end{figure}

Object reconstruction performance faces unique challenges, as highlighted by the red cells. In the egocentric view, it is notably more difficult to reconstruct accurate object surfaces, articulation (AAE), and hand-object contact (CDev). This increased difficulty arises because objects are often positioned at the image edges and obscured by human arms. Additionally, object poses exhibit greater diversity in the egocentric view due to varying camera angles and occlusions.
While a static camera maintains consistent camera extrinsics across a sequence, an egocentric camera's extrinsics change with each frame, resulting in higher diversity in camera-view object 6D poses. This diversity complicates object pose estimation in egocentric views.
\refFig{fig:failure_cases} illustrates these challenges using the best-performing method, \RWTH. Despite achieving reasonable hand poses, object poses are significantly impacted by occlusions from hands and arms, and the egocentric view undergoes substantial changes throughout a sequence.

\myparagraph{Object-wise evaluation}
\refFig{fig:ego_allo_ratio}b and \ref{fig:ego_allo_ratio}c break down performances on different objects in the egocentric and allocentric view test sets. 
The best method in both cases is \RWTH. {The hardest object} to reconstruct with good contact consistency (see CDev) is the microwave in both settings due to global rotation sensitivity (see \reffig{fig:failure_cases}), though this can be mitigated by a keypoint-based approach~\cite{hampali2022keypoint}. Estimating objects in the egocentric view is more difficult than in the allocentric view, which is indicated by higher errors for all methods.

\begin{figure}[t]
  \begin{minipage}[t]{0.5\linewidth}
    \centering
    \includegraphics[width=\linewidth]{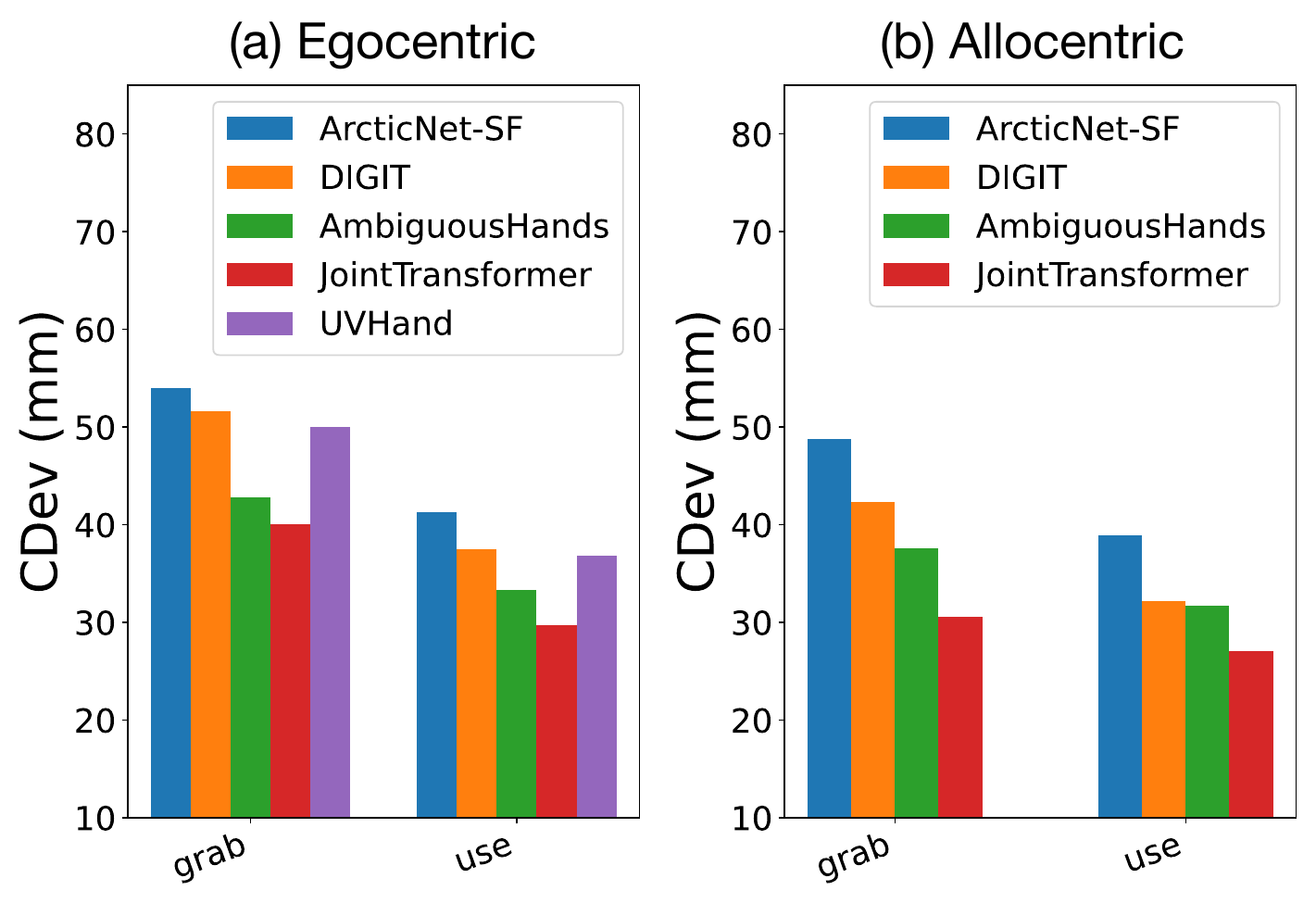}
    \caption{\mytitle{Hand-object contact quality for reconstructed results per action} 
    We evaluate the contact quality of the \threeD reconstruction results from all methods for each action (\ie, grab or use), using Contact Deviation (CDev) in mm as the metric, where lower values indicate better quality.}
    \label{fig:action_breakdown}
  \end{minipage}
  \hfill
  \begin{minipage}[t]{0.47\linewidth}
    \centering
    \includegraphics[width=\linewidth]{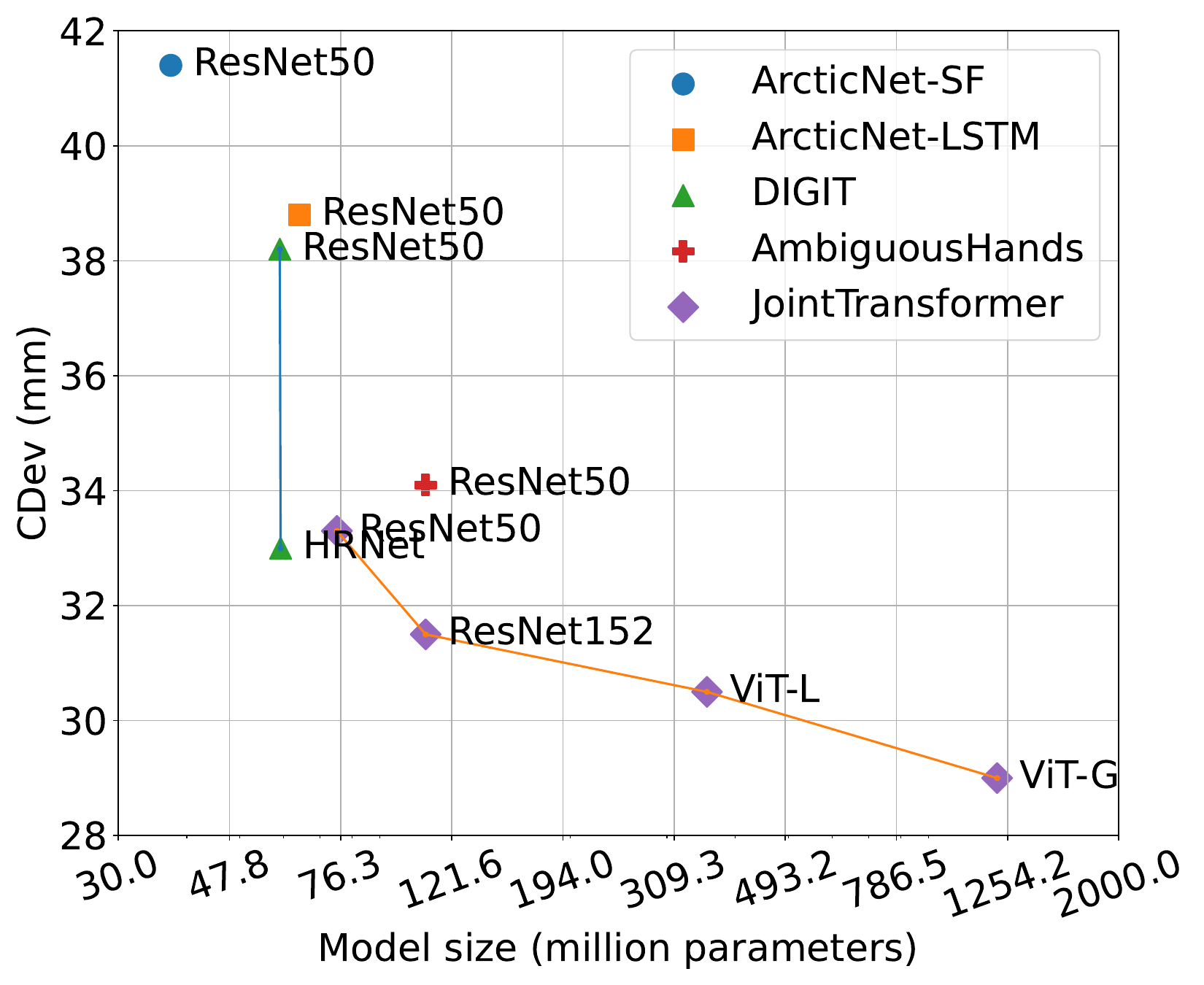}
    \caption{\mytitle{Contact deviation  vs. model size} We assess the contact quality of the reconstruction results, varying by the number of parameters in each model. Contact quality is measured using Contact Deviation (CDev) in mm, with lower values indicating superior results. }
    \label{fig:size_vs_cdev}
  \end{minipage}
\end{figure}

\myparagraph{Action-wise evaluation}
\refFig{fig:action_breakdown} compares performance of different methods in ``grab'' and ``use'' actions. In \arctic, there are sequences to interact with the object with two types of actions by either not articulating the object, or allowing object articulation. 
Interestingly, the ``grab'' motion is more challenging in egocentric and allocentric views. 
We hypothesize that this is because there are more diverse object poses for ``grab'' motions since during object articulation, the participants often focus on articulation instead of object manipulation.

\myparagraph{Effect of model size}
\refFig{fig:size_vs_cdev} illustrates the impact of model size on hand-object contact performance, measured by CDev, for reconstruction on the allocentric validation set. Most methods utilize ResNet50, with \RWTH being the top performer. As the trainable parameters in the backbone increase, \RWTH consistently reduces the CDev error. Note that the x-axis is in log-scale. \RWTH achieved a CDev error of $30.5$mm with ViT-L and $29.0$mm with ViT-G, which has ten times more parameters than ViT-L. Interestingly, \RWTH uses frozen weights in the large-scale ViT-L and ViT-G backbones, yet achieves the best results. This suggests a potential direction for leveraging large-scale foundational backbones for hand-object reconstruction.

\section{Conclusion}
In this paper, we introduce the HANDS23 challenge and provide analysis based on the results of the top submitted methods and more recent baselines on the leaderboards. 
We organize and compare the submissions and their implementation details based on the learning methods, architecture, pre- and post-processing techniques, and training configurations.
We thoroughly analyze various aspects, such as hand-object occlusions, action and object-wise evaluation, distortion correction, multi-view fusion, egocentric-allocentric comparison, and performance gain of large transformer models.

\myparagraph{Future directions}
There are several future directions that the community can take. For example, one can explore
more efficient training using multi-view egocentric cameras, leveraging 3D foundation priors~\cite{poole2022dreamfusion,liu2023zero1to3} to regularize template-free hand-object reconstruction~\cite{fan2024hold}, estimating hand-object poses with more expressive representations (\eg, heatmap-based approaches~\cite{hampali2022keypoint}),
incorporating motion and temporal modeling~\cite{fu2023deformer}, featuring more diverse egocentric interaction scenarios,
recognizing actions through captured hand poses~\cite{shamil2024utility}, 
learning robotic grasping from reconstructed hand-object pose sequences,
and so forth.

\clearpage  %

\section*{Acknowledgements}
This work is supported by the Ministry of Science and ICT, Korea, under the ITRC 
program (IITP-2024-2020-0-01789).
supervised by the IITP.
UTokyo is supported by
JST ACT-X Grant Number JPMJAX2007, JSPS KAKENHI Grant Number JP22KJ0999, 
JST Adopting Sustainable Partnerships for Innovative Research Ecosystem (ASPIRE) Grant Number JPMJAP2303.
NUS is supported by the Ministry of Education, Singapore, under its MOE Academic Research Fund Tier 2 (STEM RIE2025 MOE-T2EP20220-0015).

\bibliographystyle{splncs04}
\bibliography{main}

\clearpage
\setcounter{section}{0}
\setcounter{figure}{0}
\setcounter{table}{0}
\renewcommand{\thesection}{\Alph{section}}
\renewcommand{\thefigure}{\alph{figure}}
\renewcommand{\thetable}{\alph{table}}
\renewcommand\thesubsection{\thesection.\arabic{subsection}}

\title{
Benchmarks and Challenges in Pose Estimation
for Egocentric Hand Interactions with Objects\\
\vspace{2mm}
\textmd{--- Supplementary Material ---}} 
\titlerunning{Benchmarks and Challenge for Egocentric Hand} 
\authorrunning{Z. Fan, T. Ohkawa, L. Yang et al.} 
\author{}
\institute{}

\maketitle

\section{Additional results of \assembly}
\begin{figure*}[t]
\centering
\includegraphics[width=1\hsize]{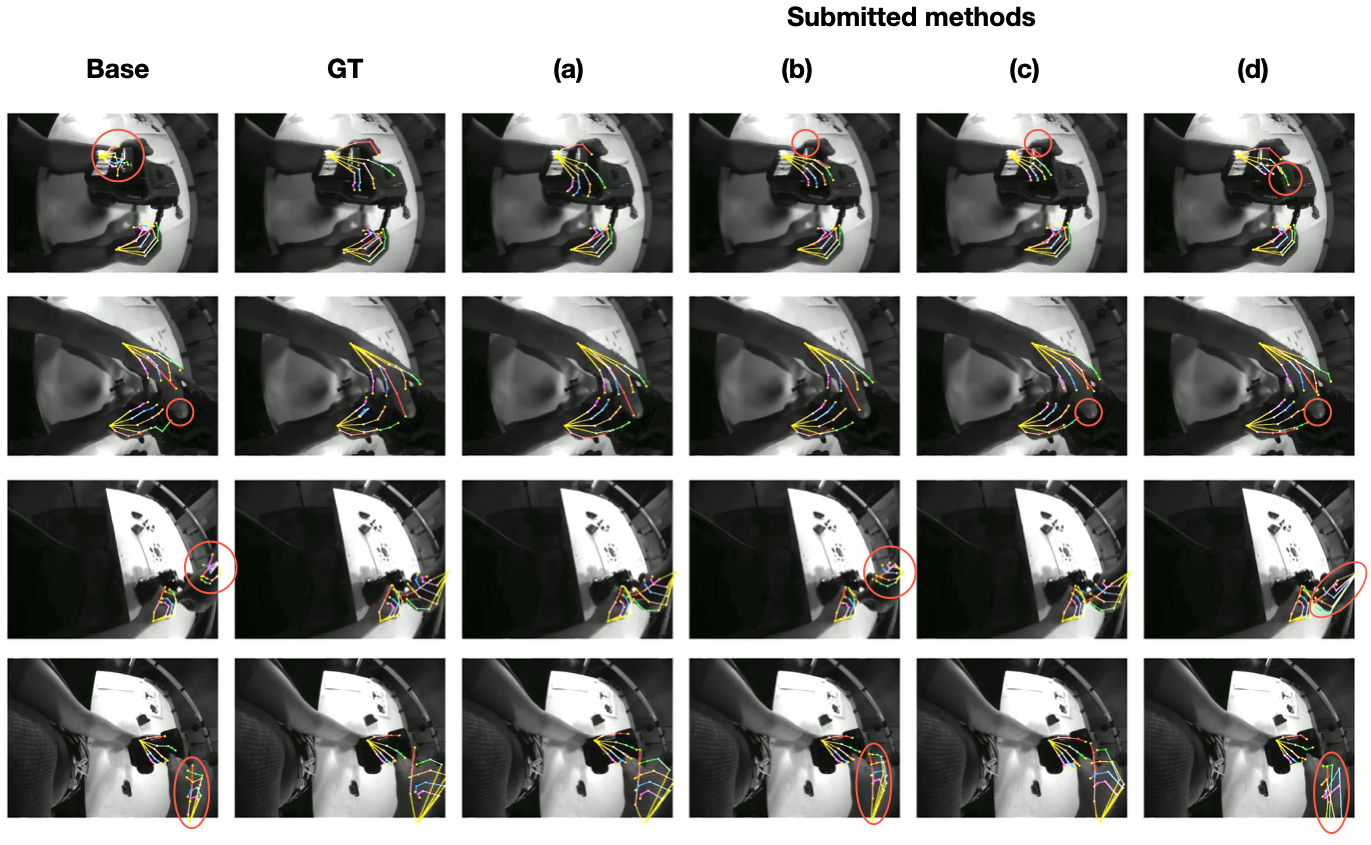}
\caption{\textbf{Qualitative results of submitted methods in AssemblyHands.}
The columns correspond to the results of \AHbase, ground-truth (GT), submitted methods, namely (a) \AHone, (b) \AHtwo, (c) \AHthree, and (d) \AHfour.
The red circles indicate where failures occur.
}
\label{fig:ah_qual}
\end{figure*}

\myparagraph{Qualitative results}
\refFig{fig:ah_qual} shows the qualitative results of submitted methods and failure patterns indicated by the red circles.
The left hand in the first row grabs the object where the left thumb finger is only visible. 
While \AHbase fails to infer the plausible pose, \AHone enables estimation in such heavy hand-object occlusions compared to the GT.
However, the methods \AHtwo and \AHthree incorrectly predict the location of the left thumb finger and \AHfour's prediction of the left index and middle fingers is also erroneous.
The second row is the case where two hands and an object are closely interacting, particularly the left thumb finger presents near the right hand.
The methods \AHbase, \AHthree, and \AHfour fail to localize the left thumb finger.
The third and fourth rows indicate hand images presented near the image edges.
The methods \AHbase, \AHtwo, and \AHfour are prone to produce implausible predictions, including noise and stretched poses due to the distortion effect discussed in Section 5.2 ``\textbf{Bias of hand position in an image}.''
The method \AHone with distortion correction successfully addresses these edge images.

\begin{figure*}[t]
\centering
\includegraphics[width=1\hsize]{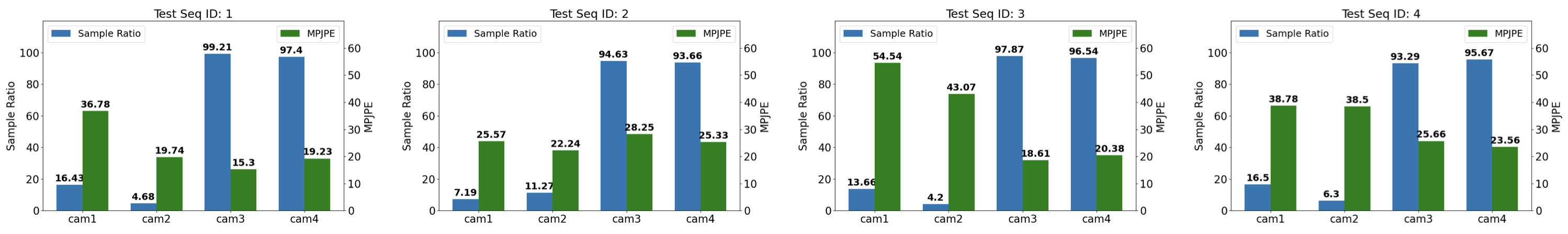}
\caption{\textbf{Additional results of multi-view fusion in AssemblyHands}.
We analyze the availability of samples and performance per camera view. 
The two lowest cameras (cam3, cam4) out of the four cameras allow us to capture hands most of the time ($>$93~\% of samples).
In contrast, the images from cam1 and cam2 are fewer and the error varies in different sequences.
}
\label{fig:ah_view}
\end{figure*}

\myparagraph{Per-view analysis}
\refFig{fig:ah_view} shows the detailed statistics and performance of per-view predictions, related to the analysis in Section 5.2 ``\textbf{Effect of multi-view fusion}.''
Considering per-sequence results, we find the sample availability (blue bars) and performance (green bars) from cam1 and cam2 vary among different users. 
In contrast, the number of samples and performance of cam3 and cam4 are mostly stable.
This study further necessitates the sample selection and multi-view fusion adaptively for each sequence (user).

\end{document}